# Explicit Dropout: Deterministic Regularization for Transformer Architectures

Vidhi Agrawal[a,*], Illia Oleksiienko[b] and Alexandros Iosifidis[c]

[a]*Faculty of Information Technology and Communications,Tampere University, Tampere, 33720, Finland*
[b]*Department of Electrical and Computer Engineering, Aarhus University, Aarhus, 8000, Denmark*
[b]*Faculty of Information Technology and Communications,Tampere University, Tampere, 33720, Finland*

ABSTRACT

Dropout is a widely used regularization technique in deep learning, but its effects are typically realized through stochastic masking rather than explicit optimization objectives. We propose a deterministic formulation that expresses dropout as an additive regularizer directly incorporated into the training loss. The framework derives explicit regularization terms for Transformer architectures, covering attention query, key, value, and feed-forward components with independently controllable strengths. This formulation removes reliance on stochastic perturbations while providing clearer and fine-grained control over regularization strength. Experiments across image classification, temporal action detection, and audio classification show that explicit dropout matches or outperforms conventional implicit methods, with consistent gains when applied to attention and feed-forward network layers. Ablation studies demonstrate stable performance and controllable regularization through regularization coefficients and dropout rates. Overall, explicit dropout offers a practical and interpretable alternative to stochastic regularization while maintaining architectural flexibility across diverse tasks.

## 1. Introduction

Regularization is a fundamental component of modern machine learning, enabling models with high expressive power to generalize beyond the training data. Classical approaches such as $\ell_2$ weight decay [19], early stopping [26], and Jacobian or gradient-norm regularization [28] directly modify the training objective to penalize complex hypotheses and sensitivity to perturbations. As deep neural networks have grown in depth, width, and structural sophistication, the design of effective regularization techniques has become increasingly critical for achieving robust generalization across diverse learning regimes.

Among existing regularization methods, dropout [15] remains one of the most widely adopted. Originally introduced as a stochastic training procedure that randomly deactivates neurons, dropout was shown to significantly reduce overfitting by preventing co-adaptation and implicitly averaging over an ensemble of subnetworks [29]. Since its introduction, dropout has been successfully applied across a wide range of architectures and tasks, including convolutional networks, recurrent models, and Transformers [21]. Subsequent work has revealed that dropout does not simply "turn off" neurons at random. It rather implicitly reshapes the loss landscape, modifies the effective hypothesis class, and induces noise to the network for more effective generalization [33, 10]. Over the past decade, a rich variety of dropout variants has been proposed, targeting different stages of the training pipeline and different structural aspects of deep models. A recent survey [21] reviews dropout variants across architectural design, embedding spaces, and input-level transformations.

Despite its empirical success, dropout is fundamentally implemented as a stochastic operation whose regularization effect is implicit, indirect, and controlled primarily through architectural choices and dropout probabilities. As a consequence, it is difficult to reason about or finely tune its regularization effect. In this work, we revisit dropout from a principled regularization perspective and ask a fundamental question: *can dropout be reformulated as an explicit, deterministic regularizer that directly controls parameter regularization strength?* Rather than treating dropout as a training-time perturbation, we seek a formulation in which its regularization effect appears explicitly as an additive term in the loss function, analogous to classical regularizers such as $\ell_2$ penalties. Such a formulation would enable precise control over regularization strength, eliminate reliance on stochastic masking, and provide clearer theoretical and practical insights into how dropout shapes learned representations. Our work makes the following contributions:

- We formulate dropout as an explicit, additive regularizer by deriving its deterministic counterpart, transforming it from a stochastic training heuristic into a principled loss-based regularization method.

- We instantiate this formulation for deep architectures and Transformer models by deriving explicit regularization terms corresponding to dropout applied on attention query, key, value, and feed-forward representations, with separate coefficients to enable fine-grained regularization strength control.

*Corresponding author

✉ vidhi.agrawal@tuni.fi (V. Agrawal); io@ece.au.dk (I. Oleksiienko); io@ece.au.dk (A. Iosifidis)

ORCID(s): 0009-0004-7594-8741 (V. Agrawal); 0000-0001-7592-365X (I. Oleksiienko); 0000-0003-4807-1345 (A. Iosifidis)

- Through extensive experiments across image classification, temporal action detection, and audio classification tasks, we show that explicit dropout consistently matches or outperforms standard implicit dropout variants.

The code for the project can be found at https://github.com/vidhi0206/Explicit-dropout.

## 2. Related Works

Dropout was introduced in [15] as a simple yet effective method to improve generalization and reduce overfitting by preventing co-adaptation. It has since become a standard component of deep neural networks, including convolutional, feed-forward, and Transformer architectures. From a theoretical perspective, dropout has been interpreted as training an implicit ensemble of subnetworks [29] or as approximate Bayesian inference under variational assumptions [8].

A complementary line of work treats dropout through complexity measures and generalization bounds. Rademacher based analyses derive bounds on the generalization gap that depend explicitly on dropout rates, revealing that the stochastic masks induce data-dependent regularizers that adapt to the learned representations [34, 1]. These results motivate adaptive dropout schemes that optimize dropout rates to minimize theoretical generalization bounds, rather than tuning them as fixed hyperparameters [34].

Beyond vanilla dropout, many works adapt the dropout rate or structure of masking. Concrete Dropout [9] introduces a continuous relaxation of Bernoulli dropout using the Concrete distribution, enabling gradient-based optimization of dropout probabilities within a Bayesian framework. Variational Nested Dropout [5] organizes features into nested subnetworks with learned importance, while Rademacher [34] dropout derives adaptive masks by optimizing generalization-gap bounds. Y-Drop [11] uses conductance based scores to drop influential neurons in fully connected layers, and Progressive Data Dropout [27] removes training samples in a curriculum manner to regularize deep face recognition. Stochastic-depth-style method [24] views layer-wise dropout as a form of depth-wise regularization that can be interpreted as a variant of stochastic depth. Augment dropout [10] separates the dropout in forward and backward passes and uses them with different rates, as forward pass dropout acts as a data augmentation and backward pass dropout acts as a process inserting noise to the model for improving generalization. It emphasized that the precise placement of dropout with respect to batch normalization and convolutions substantially affects both regularization strength and computational efficiency. [2] systematically explores ordering and insertion points of dropout and batch normalization, demonstrating that inappropriate placement can hurt convergence or waste compute, whereas carefully chosen positions yield better accuracy–efficiency trade-offs.

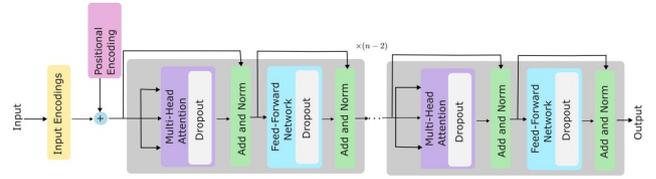

**Figure 1:** Transformer encoder architecture highlighting all locations where dropout can be applied, including within the multi-head attention mechanism and the feed-forward network.

### 2.1. Dropout in Transformer Architectures

Transformers [32] have become the dominant architecture for data modeling in language, vision, audio, and multimodal tasks, largely due to their ability to capture long-range dependencies through attention. A Transformer encoder consists of stacked layers, each composed of a multi-head attention sublayer, a feed-forward network, residual connections, layer normalization, and dropout applied throughout (Fig. 1). In a standard encoder layer, the input $\mathbf{X} \in \mathbb{R}^{N \times d}$, formed by $N$ tokens with embedding dimension $d$, is first linearly projected into query, key, and value representations:

$$\mathbf{Q} = \mathbf{X}\mathbf{W}_q^T, \quad \mathbf{K} = \mathbf{X}\mathbf{W}_k^T, \quad \mathbf{V} = \mathbf{X}\mathbf{W}_v^T, \quad (1)$$

where $\mathbf{W}_q, \mathbf{W}_k, \mathbf{W}_v \in \mathbb{R}^{d \times d}$ are trainable parameter matrices.

The scaled dot-product attention mechanism computes similarity scores stored in $\mathbf{S}$, which are normalized row-wise by a softmax to obtain attention weights $\mathbf{A}$ and the attention output:

$$\mathbf{S} = \frac{\mathbf{Q}\mathbf{K}^T}{\sqrt{d}}, \quad (2)$$
$$\mathbf{A} = \text{softmax}(\mathbf{S}), \quad \text{Attention}(\mathbf{Q}, \mathbf{K}, \mathbf{V}) = \mathbf{A}\mathbf{V}.$$

Equivalently, one may write $\text{Attention}(\mathbf{Q}, \mathbf{K}, \mathbf{V}) = \mathbf{A}\mathbf{V}\mathbf{D}^{-1}$, where $\mathbf{D} = \text{diag}(\mathbf{A}\mathbb{1}_N)$ contains the row sums of $\mathbf{A}$ and $\mathbb{1}_N$ is an $N$-dimensional vector of ones, highlighting the normalization inherent in the softmax.

Dropout is typically applied at multiple locations inside this encoder block, such as the attention block or the feed-forward network. DropAttention [23] and DropKey [20] explore the different positionings of dropout within the multi-head attention block. Subsequent works explored structured variants such as head dropout [39] and LayerDrop [7] which stochastically removes entire Transformer heads and layers respectively during training. These approaches improve robustness and generalization, particularly in large-scale models. R-Drop [22] enforces consistency between multiple stochastic forward passes with dropout, effectively tightening the output distribution.

### 2.2. Explicit Dropout Regularization

While dropout is often implemented as stochastic masking, several works make its explicit regularization effect precise by rewriting the expected dropout objective as an empirical risk plus an additive penalty term. For generalized linear models, Wager et al. [33] show that dropout training is equivalent to solving a deterministic problem with an adaptive $\ell_2$–type regularizer that scales features by an estimate of the inverse Fisher information. They further disentangle explicit and implicit effects, defining the explicit regularizer as the difference between the expected dropout objective and the standard loss. Iosifidis et al. [17] incorporate dropout into the linear regression step of single-hidden layer randomized networks and provide a closed-form formulation. They show that dropout can be interpreted as inducing an explicit weight-dependent regularizer by minimizing the discrepancy between original and noise-perturbed representations. Arora et al. [1] extend this line of work and provide explicit forms and capacity-control guarantees for dropout in matrix completion and two-layer ReLU networks, showing that dropout induces a data-dependent regularizer whose value directly controls Rademacher complexity and generalization bounds.

We focus on the explicit regularizer induced by dropout in a two-layer feed-forward network, following the formulation of Arora et al. [1]. Consider a network

$$f(\mathbf{x}; \mathbf{W}_1, \mathbf{W}_2) = \sum_{j=1}^{d_1} \sigma(\mathbf{x}\mathbf{W}_{1,j}^T) \mathbf{W}_{2,j}^T, \quad (3)$$

where $\mathbf{x} \in \mathbb{R}^d$ is the input vector, $\mathbf{W}_1 = [\mathbf{w}_{1,1}, \ldots, \mathbf{w}_{1,d_1}] \in \mathbb{R}^{d_1 \times d}$ is the first-layer weight matrix, $\mathbf{W}_2 \in \mathbb{R}^{d_2 \times d_1}$ is the second-layer weight matrix, and $\sigma(\cdot)$ is the ReLU activation function. Let $\{(\mathbf{x}_i, y_i)\}_{i=1}^N$ be the training set, and let $\mathbf{B} = \mathrm{diag}(B_{11}, \ldots, B_{d_1 d_1})$ be a diagonal random matrix with

$$B_{jj} \sim \frac{1}{1-p} \text{Bernoulli}(1-p), \quad j \in [d_1], \quad (4)$$

for dropout rate $p \in (0, 1)$. The diagonal matrix represents column dropout, i.e., dropout of certain feature embeddings from the whole input mini-batch. Arora et al. [1] show that the final loss function can be decomposed as

$$L(\mathbf{W}_1, \mathbf{W}_2) = \frac{1}{n} \sum_{i=1}^N \left(y_i - f(\mathbf{x}_i; \mathbf{W}_1, \mathbf{W}_2)\right)^2 + \widehat{R}(\mathbf{W}_1, \mathbf{W}_2), \quad (5)$$

where $\widehat{R}(\mathbf{W}_1, \mathbf{W}_2)$ is the *explicit* regularizer due to dropout. For the two-layer network above with ReLU activations, the explicit regularizer takes the form

$$\widehat{R}(\mathbf{W}_1, \mathbf{W}_2) = \lambda \sum_{j=1}^{d_1} \|\mathbf{w}_{2,j}\|_2^2 \, \widehat{a}_j^2, \quad (6)$$

$$\lambda = \sqrt{\frac{p}{1-p}}, \quad \widehat{a}_j^2 = \frac{1}{n} \sum_{i=1}^N \sigma(\mathbf{x}_i \mathbf{w}_{1,j}^T)^2.$$

$\widehat{a}_j^2$ is the empirical second moment of the $j$-th hidden neuron's activation. Thus, dropout induces a data-dependent penalty that couples the norm of each hidden weight $w_{1,j}$ with the magnitude of its responses on the training data. Increasing the dropout rate $p$ increases $\lambda = \sqrt{p/(1-p)}$, thereby intensifying the penalty and tightening generalization bounds in terms of the value of $\widehat{R}(\mathbf{W}_1, \mathbf{W}_2)$.

While this explicit dropout formulation provides useful theoretical insight, it has limitations when extended to modern attention-based architectures. It assumes columnwise (feature-level) dropout, with a regularizer acting independently per feature, failing to capture cross-feature interactions that are central to attention mechanisms based on matrix multiplications and softmax normalization. These structural considerations motivate the development of an explicit regularizer tailored to Transformer architectures, which we describe in the following section. Section B of Appendix provides the relation and highlights the differences between the regularizer presented by Arora et al. [1] and the proposed regularizer.

## 3. Explicit Transformer Dropout Regularization

We propose an explicit regularization framework for Transformer architectures by formulating dropout on the attention query, key, value, and feed-forward representations as an additive regularization term in the training objective. Unlike conventional dropout, which is applied implicitly through stochastic masking, our formulation yields a deterministic regularizer analogous to $\ell_2$ regularizer that can be directly added to the task loss, and its strength can be balanced using a regularization coefficient $\lambda$.

### 3.1. Dropout as a Regularizer on Queries

We derive the regularization loss induced by applying dropout to the query representations, expressing a weight-dependent regularizer [17] to attention queries. Specifically, we consider standard inverted dropout, where a Bernoulli mask $\mathbf{M}_t = [\mathbf{m}_{1t}, \mathbf{m}_{2t}, \ldots, \mathbf{m}_{Nt}] \in \{0, 1\}^{N \times d}$ with keep probability $(1 - p)$ is applied to the query vectors across training epochs indexed by $t$.

Let $\mathbf{X} = [\mathbf{x}_1, \mathbf{x}_2, \ldots, \mathbf{x}_N] \in \mathbb{R}^{N \times d}$ denote the input tokens, where each $\mathbf{x}_i \in \mathbb{R}^{1 \times d}$. For a query corresponding to token $\mathbf{x}_i$, the attention score is defined as

$$\mathbf{S}_i = \mathbf{x}_i \mathbf{W}_q^T \mathbf{W}_k \mathbf{X}^T. \quad (7)$$

During training, dropout is applied to the input representation at epoch $t$ as $\mathbf{x}_{it} = \mathbf{m}_{it} \odot \mathbf{x}_i$, where $\odot$ denotes element-wise multiplication. The resulting attention scores become

$$\mathbf{S}_{it} = (\mathbf{m}_{it} \odot \mathbf{x}_i) \mathbf{W}_q^T \mathbf{W}_k \mathbf{X}^T. \quad (8)$$

To encourage consistency between the original and dropout-perturbed queries over training, we seek weights that minimize the discrepancy between $\mathbf{S}_i$ and $\mathbf{S}_{it}$:

$$\mathbf{S}_i - \mathbf{S}_{it} = \tilde{\mathbf{x}}_{it} \mathbf{W}_q^T \mathbf{W}_k \mathbf{X}^T, \quad (9)$$

where $\tilde{\mathbf{x}}_i = \mathbf{x}_i - (\mathbf{m}_{it} \odot \mathbf{x}_i)$.

Consider $N_T$ stochastic training epochs with dropout applied to the input $\mathbf{X}$, leading to $\tilde{\mathbf{X}}_t = [\tilde{\mathbf{x}}_{1t}, \tilde{\mathbf{x}}_{2t}, \ldots, \tilde{\mathbf{x}}_{Nt}] \in \mathbb{R}^{N \times d}$ representing the deviations between the original and the dropout-perturbed representations at epoch $t$. The proposed regularizer is defined as:

$$J_q = \frac{1}{2N_T} \sum_{t=1}^{N_T} \left\| \tilde{\mathbf{X}}_t \mathbf{W}_q^T \mathbf{W}_k \mathbf{X}^T \right\|_F^2. \quad (10)$$

Using $\|\mathbf{A}\|_F^2 = \mathrm{Tr}(\mathbf{A}^T \mathbf{A})$ and rearranging terms

$$J_q = \frac{1}{2} \mathrm{Tr}\left[ \left( \frac{1}{N_T} \sum_{t=1}^{N_T} \tilde{\mathbf{X}}_t^T \tilde{\mathbf{X}}_t \right) \mathbf{W}_q^T \mathbf{W}_k \mathbf{X}^T \mathbf{X} \mathbf{W}_k^T \mathbf{W}_q \right]. \quad (11)$$

Let

$$\mathbf{B} = \frac{1}{N_T} \sum_{t=1}^{N_T} \tilde{\mathbf{X}}_t^T \tilde{\mathbf{X}}_t, \quad \Lambda_q = \mathbf{W}_q^T \mathbf{W}_k \mathbf{X}^T \mathbf{X} \mathbf{W}_k^T \mathbf{W}_q. \quad (12)$$

Assuming independent Bernoulli dropout with drop probability $p$, the expectation over masks yields

$$\mathbf{B} \approx (\mathbf{X}^T \mathbf{X}) \odot p^2. \quad (13)$$

The resulting query regularizer becomes

$$J_q = \frac{1}{2} \mathrm{Tr}\left[ \left( (\mathbf{X}^T \mathbf{X}) \odot p^2 \right) \Lambda_q \right]. \quad (14)$$

Full derivation of $\mathbf{B}$ is provided in Section A of the Appendix.

### 3.2. Key and Value Regularizers

*Key Regularization.* We follow a similar analysis for the attention keys, yielding the following regularizer:

$$J_k = \frac{1}{2} \mathrm{Tr}\left[ \left( (\mathbf{X}^T \mathbf{X}) \odot p^2 \right) \Lambda_k \right], \quad (15)$$
$$\Lambda_k = \mathbf{W}_k^T \mathbf{W}_q \mathbf{X}^T \mathbf{X} \mathbf{W}_q^T \mathbf{W}_k.$$

*Value Regularization (Token-level Dropout).* When dropout is applied directly to the input tokens before value projection, the induced regularizer takes the form:

$$J_v = \frac{1}{2} \mathrm{Tr}\left[ \left( (\mathbf{X}^T \mathbf{X}) \odot p^2 \right) \Lambda_v \right], \quad (16)$$
$$\Lambda_v = \mathbf{W}_v^T \mathbf{W}_v.$$

This formulation reflects the effect of feature-level noise propagation through the value projection, leading to a penalty on the value transformation.

*Value Regularization (Attention-conditioned Dropout).* When dropout is instead applied after attention mixing, the resulting regularizer becomes:

$$J_{av} = \frac{1}{2} \mathrm{Tr}\left[ \left( (\mathbf{X}^T \mathbf{A}^T \mathbf{A} \mathbf{X}) \odot p^2 \right) \Lambda_v \right], \quad (17)$$

$$\Lambda_v = \mathbf{W}_v^T \mathbf{W}_v.$$

Here, the attention matrix $\mathbf{A}$ modulates the effective structure, making the regularization explicitly dependent on the learned attention patterns.

Detailed derivations of the above regularizers are provided in Section A of the Appendix.

### 3.3. Feed-Forward Network Regularizer

Each Transformer layer contains a position-wise feed-forward network

$$\mathrm{FF}(X) = \sigma(X \mathbf{W}_{\mathrm{ff},1}) \mathbf{W}_{\mathrm{ff},2}, \quad (18)$$

where $\mathbf{W}_{\mathrm{ff},1}$ and $\mathbf{W}_{\mathrm{ff},2}$ denote the first- and second-layer weight matrices, respectively. Following the derivation used for the attention projections, dropout induces an explicit regularizer on both feed-forward weight matrices. The resulting regularization terms are:

$$J_{\mathrm{ff},m} = \frac{1}{2} \mathrm{Tr}\left[ \left( (\mathbf{X}^T \mathbf{X}) \odot p^2 \right) \Lambda_{\mathrm{ff},m} \right], \quad (19)$$
$$\Lambda_{\mathrm{ff},m} = \mathbf{W}_{\mathrm{ff},m}^T \mathbf{W}_{\mathrm{ff},m}$$

where $m \in \{1, 2\}$ denotes the corresponding feed-forward network layer. Thus, dropout in the feed-forward network leads to a quadratic, data-dependent regularization term on each of the two linear transformations. The detailed derivation is provided in Section A of the Appendix.

### 3.4. Final Training Objective

The final training objective combines the task-specific loss with the dropout-induced regularizers from all Transformer components. Let $L$ denote the number of encoder blocks. The overall objective is defined as

$$J_{\mathrm{final}} = J_{\mathrm{task}} + J_q + J_k + J_v + J_{\mathrm{ff}}. \quad (20)$$

Each term aggregates the corresponding regularizers across layers. For example, the query regularizer is given by

$$J_q = \sum_{l=1}^{L} \lambda_q^{(l)} J_q^{(l)}, \qquad \lambda_q^{(l)} \geq 0, \quad (21)$$

where $J_q^{(l)}$ denotes the query regularizer at layer $l$, and $\lambda_q^{(l)}$ controls its strength. Setting $\lambda_q^{(l)} = 0$ effectively disables the regularizer for that layer. Analogous expressions hold for the key, value, and feed-forward network regularizers:

$$J_k = \sum_{l=1}^{L} \lambda_k^{(l)} J_k^{(l)}, \quad (22)$$

$$J_v = \sum_{l=1}^{L} \lambda_v^{(l)} J_v^{(l)},$$

$$J_{\mathrm{ff}} = \sum_{l=1}^{L} \sum_{m=1}^{2} \lambda_{\mathrm{ff}}^{(l)} J_{\mathrm{ff},m}^{(l)},$$

with layer-wise regularization coefficients $\lambda_k^{(l)}, \lambda_v^{(l)}, \lambda_{\mathrm{ff}}^{(l)} \geq 0$. This formulation allows independent control of the regularization strength for each component and each layer of the Transformer encoder.

## Table 1
Effect of implicit and explicit dropout regularization on CIFAR-10 with dropout ratio of 0.2 using a 7-layer Vision Transformer. Results report test accuracy.

| Dropout on Attention Sequence | Dropout on FF network | Accuracy (%) |
|---|---|---|
| DropAttention (Implicit) [23] | Implicit | 85.24 ± 0.75 |
| DropKey (Implicit) [20] | Implicit | 85.45 ± 0.41 |
| None | Implicit | 85.14 ± 0.63 |
| (Q) Arora et al. [1] | Explicit | 61.08 ± 0.80 |
| (K) Arora et al. [1] | Explicit | 47.01 ± 1.58 |
| (V) Arora et al. [1] | Explicit | 59.22 ± 1.80 |
| none | Arora et al. [1] | 86.02 ± 0.51 |
| None | Explicit | 86.14 ± 0.36 |
| Explicit (Q) | Explicit | 83.92 ± 0.98 |
| Explicit (K) | Explicit | 84.06 ± 0.85 |
| Explicit (V) | Explicit | **86.38 ± 0.44** |
| Explicit (AV) | Explicit | 86.11 ± 0.41 |

## 4. Experiments

We evaluate the proposed explicit dropout regularization across diverse tasks and Transformer architectures. Our goal is to assess whether formulating dropout as an explicit, additive regularizer can match or even exceed the performance of commonly used implicit dropout strategies, including DropAttention [23] and DropKey [20], while providing more direct control over regularization strength. In addition, we also adapt prior explicit dropout formulations for Transformers and perform direct comparisons. This allows us to evaluate whether our formulation offers practical improvements in terms of performance. We conduct experiments on image classification, temporal action detection, and audio classification benchmarks, covering both vision-only Transformers and Transformer encoders operating on pretrained features. Unless otherwise stated, all methods use identical training settings for fair comparison.

### 4.1. Experimental Setup
#### 4.1.1. Datasets and Baselines:

For image classification, experiments are conducted on CIFAR-10 and CIFAR-100 [18]. For both datasets, the original training set is further split into training and validation subsets using a 70:30 ratio, while the standard test set is used for final evaluation. For temporal action detection, we report results on THUMOS14 [16], which contains 413 training videos spanning 20 action categories with frame-level annotations. Similar to the CIFAR setup, the provided training set is split into training and validation subsets in a 70:30 ratio, and evaluation is performed on the official test set. Following prior work [36], we employ a Temporal Segment Network (TSN) [35] pretrained on ActivityNet [14] or Kinetics-400 [3] for feature extraction, producing sequences of 64 tokens. For audio classification, we utilize the GTZAN Music Genre Classification dataset [31]. Consistent with previous studies [13, 37], Mel spectrograms are extracted and converted into sequences of 120 tokens using a VGGish network, after which the data are split into training, validation, and test sets.

Our primary baseline consists of Transformer encoder layers [32] trained with conventional dropout applied to attention modules or feed-forward network layers. We compare against related stochastic attention strategies, including DropKey [20] and DropAttention [23], as well as variants with dropout disabled in certain modules or applied explicitly through formulations from [1].

#### 4.1.2. Models Used:

For CIFAR-10 and CIFAR-100, we use a Vision Transformer (ViT) [6, 25] with seven Transformer encoder layers and a patch-based embedding and classification token (CLS). For THUMOS14 and GTZAN, we employ a lightweight Transformer encoder with two layers [12] operating on pre-extracted video and audio features, respectively. Across all models, we evaluate explicit dropout regularization applied to Query (Q), Key (K), Value (V), and feed-forward network layers, controlled by the regularization coefficients $\lambda$. Experiments are conducted across multiple values of the regularization coefficient $\lambda$ and learning rates. For each experiment, the reported test results correspond to the configuration achieving the highest validation accuracy. This protocol ensures a fair comparison between implicit and explicit regularization strategies.

### 4.2. Evaluation Results

Tables 1–4 present a comprehensive comparison between implicit dropout strategies and the proposed explicit regularization across multiple modalities and model scales. Overall, the results demonstrate that explicit dropout yields competitive or superior performance compared to conventional implicit methods while offering more controllable regularization behavior. For all experiments, we report the mean and standard deviation of the evaluation metrics over five independent runs with different random seeds. This

**Table 2**
Effect of implicit and explicit dropout regularization on CIFAR-100 with dropout ratio of 0.2 using a 7-layer Vision Transformer. Results report test accuracy.

| Dropout on Attention Sequence | Dropout on FF network | Accuracy (%) |
|---|---|---|
| DropAttention (Implicit) [23] | Implicit | 58.01 ± 1.20 |
| DropKey (Implicit) [20] | Implicit | **59.11 ± 1.23** |
| None | Implicit | 59.01 ± 1.13 |
| (Q) Arora et al. [1] | Explicit | 37.59 ± 0.79 |
| (K) Arora et al. [1] | Explicit | 29.61 ± 0.85 |
| (V) Arora et al. [1] | Explicit | 37.16 ± 1.51 |
| none | Arora et al. [1] | 55.42 ± 1.37 |
| None | Explicit | 56.81 ± 2.84 |
| Explicit (Q) | Explicit | 53.79 ± 4.82 |
| Explicit (K) | Explicit | 51.53 ± 0.90 |
| Explicit (V) | Explicit | 55.15 ± 0.84 |
| Explicit (AV) | Explicit | 56.62 ± 2.03 |

protocol ensures a fair comparison and provides a reliable estimate of performance stability.

*CIFAR-10 and CIFAR-100.* On CIFAR-10 (Table 1), explicit dropout—particularly when applied to the value (V) branch—achieves the best overall performance, reaching 86.38%, outperforming all implicit methods and other explicit variants. While implicit approaches such as DropAttention and DropKey remain in a competitive range, they do not surpass the strongest explicit configuration. In contrast, Arora et al. [1] is competitive only when applied to the feed-forward network, while its performance drops significantly when applied to attention components. Overall, CIFAR-10 results indicate that value-based explicit regularization provides the most effective inductive bias for improving generalization.

On CIFAR-100 (Table 2), the task is more sensitive to the choice of regularization strategy. Among implicit methods, DropKey performs best at 59.11%, slightly outperforming other implicit variants. Explicit dropout remains competitive but is generally weaker overall, with the best explicit configuration (AV) reaching 56.62%. Despite this gap, explicit methods still provide stable performance and offer a more structured and controllable alternative to implicit dropout across settings.

*THUMOS14 (Kinetics and ActivityNet Features).* For THUMOS14 features extracted from TSN backbones pre-trained on Kinetics and ActivityNet, explicit dropout again demonstrates strong performance. On Kinetics features, Explicit (V)/Explicit achieves the highest mAP (64.68%), slightly surpassing both implicit DropKey and the None/Explicit baseline. A similar trend holds for ActivityNet features, where Explicit (V)/Explicit reaches 56.51%, outperforming all implicit configurations. In contrast, the Arora et al. [1]-style attention dropout variants exhibit larger variance and weaker performance under similar dropout ratios. This suggests that our formulation improves robustness in temporally structured representations where attention dynamics are more complex.

*GTZAN (Audio Classification).* On the GTZAN audio classification benchmark (Table 4), implicit dropout methods already achieve strong performance ($\approx$ 85% accuracy). The performance differences between methods are comparatively smaller, reflecting the relatively simpler structure of the dataset. Nevertheless, the proposed dropout remains competitive (85.78% accuracy), with explicit (K) achieving the best overall accuracy. This indicates that key-based regularization can be beneficial in low-data or low-complexity regimes, where over-regularization may otherwise degrade performance.

Across datasets and modalities, explicit regularization applied to values, or feed-forward network components consistently provides strong or state-of-the-art performance, while explicit query and key dropout demonstrate task-dependent sensitivity. The regularization from Arora et al. [1] consistently performs poorly when applied to attention input values, i.e., Q, K, and V, as the regularizer is based on dropping out whole specific features instead of some particular dimensions in the features. This causes instability in the attention output, therefore not allowing the model to learn the patterns fully. The results indicate that explicit dropout can match or exceed implicit methods, particularly on complex datasets and structured sequence tasks, supporting the effectiveness and generality of the proposed framework.

### 4.3. Ablation Study

We conduct ablation experiments on CIFAR-10 to analyze the effect of explicit dropout regularization across different attention components and hyperparameter settings. In particular, we evaluate explicit dropout applied independently to the key (K), query (Q), and value (V) projections while varying both the learning rate and the regularization coefficient $\lambda_{dr}$. Results are summarized in Table 5.

From Table 5, we observe distinct trends in how explicit dropout interacts with different attention components. Applying dropout to the value (V) projection consistently yields the strongest performance across most hyperparameter configurations, with accuracy peaking around 86.4% at a moderate dropout regularization weight ($\lambda = 0.0005$) and learning rate 0.0005. This suggests that injecting stochasticity into the value representation helps improve robustness without destabilizing attention distributions.

On the other hand, dropout applied to the key (K) or query (Q) projections leads to noticeably higher variance and degraded accuracy, especially at larger learning rates (e.g., 0.005). This degradation likely stems from the fact that perturbations in K and Q directly affect the attention weighting mechanism, leading to unstable gradients and noisier optimization.

A combined dropout on the attention–value (AV) pathways produces intermediate results, closer to those of V-only dropout, reinforcing that controlling noise primarily in the value stream retains the most beneficial regularization effect.

**Table 3**
Comparison of implicit and explicit dropout on THUMOS14 features, extracted from the TSN model pre-trained on ActivityNet and Kinetics-400, with dropout ratio of 0.2 using a 2-layer Transformer encoder. Results report test mAP.

| Dropout on Attention Sequence | Dropout on FF network | Kinetics | ActivityNet |
|---|---|---|---|
| DropAttention [23] | Implicit | 62.03 ± 0.26 | 53.89 ± 0.46 |
| DropKey [20] | Implicit | 64.58 ± 0.33 | 55.87 ± 0.61 |
| None | Implicit | 64.53 ± 0.47 | 55.62 ± 0.74 |
| (Q) Arora et al. [1] | Explicit | 48.66 ± 6.52 | 46.67 ± 2.03 |
| (K) Arora et al. [1] | Explicit | 51.84 ± 4.14 | 44.52 ± 5.80 |
| (V) Arora et al. [1] | Explicit | 52.23 ± 7.40 | 36.96 ± 2.28 |
| None | Arora et al. [1] | 57.76 ± 0.90 | 48.44 ± 1.05 |
| None | Explicit | 63.96 ± 0.46 | 54.92 ± 0.33 |
| Explicit (Q) | Explicit | 63.70 ± 0.56 | 54.98 ± 0.84 |
| Explicit (K) | Explicit | 63.78 ± 0.54 | 54.92 ± 0.76 |
| Explicit (V) | Explicit | **64.68 ± 0.36** | **56.51 ± 0.37** |
| Explicit (AV) | Explicit | 64.24 ± 0.40 | 56.10 ± 0.55 |

**Table 4**
Comparison of implicit and explicit dropout on GTZAN (audio classification dataset) with dropout ratio of 0.2 using a 2-layer Transformer encoder. Results report test accuracy.

| Dropout on Attention Sequence | Dropout on FF network | Accuracy (%) |
|---|---|---|
| DropAttention [23] | Implicit | 84.84 ± 0.89 |
| DropKey [20] | Implicit | 84.84 ± 0.89 |
| None | Implicit | 85.00 ± 0.86 |
| (Q) Arora et al. [1] | Explicit | 85.31 ± 0.86 |
| (K) Arora et al. [1] | Explicit | 84.22 ± 1.02 |
| (V) Arora et al. [1] | Explicit | 85.16 ± 0.55 |
| None | Arora et al. [1] | 84.84 ± 0.43 |
| None | Explicit | 85.16 ± 0.55 |
| Explicit (Q) | Explicit | 85.68 ± 0.65 |
| Explicit (K) | Explicit | **85.78 ± 0.65** |
| Explicit (V) | Explicit | 85.00 ± 0.86 |
| Explicit (AV) | Explicit | 85.00 ± 0.35 |

Overall, these results highlight that the placement of dropout within the attention mechanism critically governs training stability. Moderate dropout regularization on the value projection strikes the right balance between representation smoothing and information retention, improving generalization on CIFAR-10 without sacrificing convergence speed.

The full set of ablation results for the remaining datasets is reported in Section 3 of the supplementary reader.

## 5. Conclusions

This work introduced an explicit formulation of dropout as an additive regularization mechanism for Transformer architectures, enabling direct and interpretable control over regularization strength across attention and feed-forward network components. Even though we did not identify a straightforward way to derive a unified regularization expression when dropout is applied simultaneously to multiple input matrices within the attention block, explicit dropout regularization can be induced to multiple architectural components by incorporating the corresponding regularization terms into the final loss.

Extensive experiments across image classification, temporal action detection, and audio classification demonstrate that explicit dropout achieves performance comparable to, or exceeding, conventional implicit dropout strategies such as DropAttention and DropKey. In particular, explicit regularization applied to value projections and feed-forward network layers consistently yields strong and stable improvements across modalities, while explicit value-based regularization proves especially beneficial in more complex classification tasks such as CIFAR-100. Results on THUMOS14 further show that explicit value dropout improves temporal modeling performance, achieving the highest mean Average Precision across multiple feature extractors. The ablation study confirms that the proposed formulation provides fine-grained control over regularization strength through the choice of the regularization coefficients and dropout ratio, allowing practitioners to balance generalization and regularization capacity.

Although derived for the processing blocks used in Transformer architectures, the proposed explicit dropout regularization naturally extends to convolutional operations. A convolution operation can be rewritten as a vector-based affine transformation by extracting local patches from the input feature maps, flattening them into vectors, and reshaping the convolutional kernels into an equivalent weight matrix $\mathbf{W}_{\text{conv}}$ while preserving structured weight sharing [4]. The corresponding explicit dropout regularizer has the same structure as the one derived for the feed-forward network

**Table 5**
Accuracy (acc) on CIFAR-10 for dropout applied on different attention matrices (K, Q, V) across regularization coefficients $\lambda$ and learning rates.

|   | Learning Rate | $\lambda$ | | | |
|---|---|---|---|---|---|
|   |   | 0.001 | 0.005 | 0.0001 | 0.0005 |
| K | 0.001 | 83.65 ± 0.81 | 79.56 ± 5.18 | 81.15 ± 1.53 | 84.06 ± 0.85 |
| K | 0.005 | 66.48 ± 2.19 | 61.02 ± 5.84 | 69.46 ± 0.94 | 68.30 ± 0.71 |
| K | 0.0001 | 78.02 ± 0.72 | 77.61 ± 0.65 | 78.80 ± 0.98 | 78.09 ± 1.01 |
| K | 0.0005 | 83.91 ± 0.84 | 81.87 ± 0.99 | 82.96 ± 1.37 | 83.73 ± 0.66 |
| Q | 0.001 | 83.44 ± 0.94 | 79.08 ± 5.31 | 80.76 ± 1.30 | 83.74 ± 0.45 |
| Q | 0.005 | 67.17 ± 0.36 | 58.33 ± 1.66 | 69.16 ± 0.72 | 68.81 ± 0.32 |
| Q | 0.0001 | 78.18 ± 0.62 | 77.65 ± 0.56 | 78.81 ± 1.03 | 78.23 ± 0.90 |
| Q | 0.0005 | 83.92 ± 0.98 | 81.93 ± 1.02 | 83.33 ± 1.22 | 83.70 ± 0.65 |
| V | 0.001 | 86.11 ± 0.38 | 85.91 ± 0.78 | 85.66 ± 0.82 | 85.85 ± 0.42 |
| V | 0.005 | 70.57 ± 1.38 | 71.31 ± 0.44 | 70.61 ± 1.44 | 70.57 ± 1.25 |
| V | 0.0001 | 78.79 ± 0.54 | 78.14 ± 0.54 | 79.28 ± 0.53 | 78.99 ± 0.51 |
| V | 0.0005 | 86.38 ± 0.44 | 86.16 ± 0.48 | 85.92 ± 0.62 | 86.05 ± 0.57 |
| AV | 0.001 | 85.47 ± 0.63 | 80.90 ± 1.63 | 85.71 ± 0.52 | 85.48 ± 0.87 |
| AV | 0.005 | 70.58 ± 0.80 | 70.23 ± 0.96 | 70.73 ± 0.78 | 70.41 ± 0.88 |
| AV | 0.0001 | 79.80 ± 0.57 | 78.88 ± 0.49 | 80.11 ± 0.41 | 79.84 ± 0.67 |
| AV | 0.0005 | 85.20 ± 0.72 | 84.30 ± 0.67 | 86.11 ± 0.41 | 86.04 ± 0.61 |

layers in Section 3.3 and is provided in Section 1 of the supplementary material.

## 6. Declaration of generative AI and AI-assisted technologies in the manuscript preparation process

During the preparation of this work, the author used ChatGPT and Perplexity AI to assist with formatting the manuscript. All generated content was critically evaluated, verified, and edited by the author, who takes full responsibility for the final content of the published article.

## A. Full Derivation of Dropout-Induced Regularizers

### A.1. Derivation of Query Regularizer

We consider $N_T$ stochastic training epochs with dropout masks applied to the input matrix $\mathbf{X}$. The query regularizer is

$$J_q = \frac{1}{2N_T} \sum_{t=1}^{N_T} \| \tilde{\mathbf{X}}_t \mathbf{W}_q^T \mathbf{W}_k \mathbf{X}^T \|_F^2. \quad (23)$$

Using $\| \mathbf{A} \|_F^2 = \mathrm{Tr}(\mathbf{A}^T \mathbf{A})$ it becomes

$$J_q = \frac{1}{2N_T} \sum_{t=1}^{N_T} \mathrm{Tr} \left[ \mathbf{X} \mathbf{W}_k^T \mathbf{W}_q \tilde{\mathbf{X}}_t^T \tilde{\mathbf{X}}_t \mathbf{W}_q^T \mathbf{W}_k \mathbf{X}^T \right]. \quad (24)$$

Rearranging the terms:

$$J_q = \frac{1}{2} \mathrm{Tr} \left[ \left( \frac{1}{N_T} \sum_{t=1}^{N_T} \tilde{\mathbf{X}}_t^T \tilde{\mathbf{X}}_t \right) \mathbf{W}_q^T \mathbf{W}_k \mathbf{X}^T \mathbf{X} \mathbf{W}_k^T \mathbf{W}_q \right]. \quad (25)$$

We define

$$\mathbf{B} = \frac{1}{N_T} \sum_{t=1}^{N_T} \tilde{\mathbf{X}}_t^T \tilde{\mathbf{X}}_t. \quad (26)$$

### A.2. Expansion of B

Let dropout masks for epoch $t$ be $\mathbf{M}_t \in \{0,1\}^{n \times d}$. Each element of the mask is sampled independently from a Bernoulli distribution with keep probability $(1-p)$. Then, at epoch $t$:

$$\tilde{\mathbf{X}}_t = \mathbf{X} - (\mathbf{M}_t \odot \mathbf{X}). \quad (27)$$

Expanding Eq 26,

$$\mathbf{B} = \frac{1}{N_T} \sum_{t=1}^{N_T} [(\mathbf{X} - \mathbf{X}_t)^T (\mathbf{X} - \mathbf{X}_t)] \quad (28)$$

$$= \frac{1}{N_T} \sum_{t=1}^{N_T} [\mathbf{X}^T \mathbf{X} - \mathbf{X}^T \mathbf{X}_t - \mathbf{X}_t^T \mathbf{X} + \mathbf{X}_t^T \mathbf{X}_t]. \quad (29)$$

Substituting $\mathbf{X}_t = \mathbf{M}_t \odot \mathbf{X}$,

$$\mathbf{B} = \frac{1}{N_T} \sum_{t=1}^{N_T} [\mathbf{X}^T \mathbf{X} - \mathbf{X}^T (\mathbf{M}_t \odot \mathbf{X}) - \quad (30)$$
$$(\mathbf{M}_t \odot \mathbf{X})^T \mathbf{X} + (\mathbf{M}_t \odot \mathbf{X})^T (\mathbf{M}_t \odot \mathbf{X})].$$

Following an element-wise analysis, for the $(j,k)$ entry of $(\mathbf{M}_t \odot \mathbf{X})^T (\mathbf{M}_t \odot \mathbf{X}) - \mathbf{X}^T (\mathbf{M}_t \odot \mathbf{X}) - (\mathbf{M}_t \odot \mathbf{X})^T \mathbf{X}$,

$$\sum_{i=1}^n \left[ (m_{t,ij} X_{ij})(m_{t,ik} X_{ik}) - X_{ij}(m_{t,ik} X_{ik}) - (m_{t,ij} X_{ij}) X_{ik} \right]. \quad (31)$$

Thus,

$$B_{jk} = \frac{1}{N_T} \sum_{t=1}^{N_T} \Big[ \sum_{i=1}^n X_{ij} X_{ik} + \quad (32)$$

$$\sum_{i=1}^{n} X_{ij}X_{ik}(m_{t,ij}m_{t,ik} - m_{t,ik} - m_{t,ij})\Big]$$

$$B_{jk} = \frac{1}{N_T}\sum_{t=1}^{N_T}\Big[\sum_{i=1}^{n}(X_{ij}X_{ik})(1 \qquad (33)$$
$$+ m_{t,ij}m_{t,ik} - m_{t,ik} - m_{t,ij})\Big]$$

$$B_{jk} = \sum_{i=1}^{n} X_{ij}X_{ik}\Big[\frac{1}{N_T}\sum_{t=1}^{N_T}(1 \qquad (34)$$
$$+ m_{t,ij}m_{t,ik} - m_{t,ik} - m_{t,ij})\Big].$$

To compute the inner summation, we can apply an importance sampling Monte Carlo integration method in reverse. According to importance sampling Monte Carlo,

$$\int f(x)p(x)\,dx \approx \frac{1}{N_T}\sum_{t=1}^{N_T} f(x_t), \quad x_t \sim p(x). \qquad (35)$$

Applying this to our case,

$$\frac{1}{N_T}\sum_{t=1}^{N_T}(1 + m_t^1 m_t^2 - m_t^1 - m_t^2) \qquad (36)$$
$$\approx \iint (1 + m^1 m^2 - m^1 - m^2)P(m^1)P(m^2)\,dm^1 dm^2. \qquad (37)$$

Since both $m^1$ and $m^2$ are binary, the integral has only four possible values:

$$\iint \left(1 + m^1 m^2 - m^1 - m^2\right) \times P(m^1) P(m^2)\,dm^1\,dm^2 \qquad (38)$$

$$= S_1 + S_2 + S_3 + S_4 = p^2, \qquad (39)$$
$$P(m^1) = (1-p)^{m^1} p^{(1-m^1)}, \qquad (40)$$
$$P(m^2) = (1-p)^{m^2} p^{(1-m^2)}, \qquad (41)$$
$$S_1 = (1 + 1 - 1 - 1)(1-p)^2, \qquad (42)$$
$$S_2 = (1 + 0 - 1 - 0)p(1-p), \qquad (43)$$
$$S_3 = (1 + 0 - 0 - 1)(1-p)p, \qquad (44)$$
$$S_4 = (1 + 0 - 0 - 0)p^2. \qquad (45)$$

Therefore,

$$\mathbf{B} \approx (\mathbf{X}^T\mathbf{X})p^2. \qquad (46)$$

### A.3. Derivation of the Key Dropout Regularizer

For the key corresponding to token $\mathbf{x}_i$, the attention score is defined as

$$\mathbf{S}_i = \mathbf{X}\mathbf{W}_q^T\mathbf{W}_k\mathbf{x}_i^T. \qquad (47)$$

When a dropout mask is applied, the attention score becomes

$$\mathbf{S}_{it} = \mathbf{X}\mathbf{W}_q^T\mathbf{W}_k(\mathbf{m}_{it} \odot \mathbf{x}_i)^T, \qquad (48)$$

where $\mathbf{m}_{it}$ denotes the dropout mask at training epoch $t$.

Assume training proceeds for $N_T$ stochastic epochs with independently sampled dropout masks applied to the input matrix $\mathbf{X}$. The key dropout regularizer is defined as

$$J_k = \frac{1}{2N_T}\sum_{t=1}^{N_T} \|\mathbf{X}\mathbf{W}_q^T\mathbf{W}_k\tilde{\mathbf{X}}_t^T\|_F^2. \qquad (49)$$

Using $\|\mathbf{A}\|_F^2 = \text{Tr}(\mathbf{A}\mathbf{A}^T)$, we obtain

$$J_k = \frac{1}{2N_T}\sum_{t=1}^{N_T} \text{Tr}\left(\mathbf{X}\mathbf{W}_q^T\mathbf{W}_k\tilde{\mathbf{X}}_t^T\tilde{\mathbf{X}}_t\mathbf{W}_k^T\mathbf{W}_q\mathbf{X}^T\right). \qquad (50)$$

By the cyclic property of the trace,

$$J_k = \frac{1}{2N_T}\sum_{t=1}^{N_T} \text{Tr}\left(\tilde{\mathbf{X}}_t^T\tilde{\mathbf{X}}_t\mathbf{W}_k^T\mathbf{W}_q\mathbf{X}^T\mathbf{X}\mathbf{W}_q^T\mathbf{W}_k\right). \qquad (51)$$

Rearranging the summation and using the definition of $\mathbf{B}$ in Eq. (4), we obtain

$$J_k = \frac{1}{2}\text{Tr}\left(\mathbf{B}\mathbf{W}_k^T\mathbf{W}_q\mathbf{X}^T\mathbf{X}\mathbf{W}_q^T\mathbf{W}_k\right). \qquad (52)$$

### A.4. Derivation of the Value Dropout Regularizer

The value representation is obtained by

$$\mathbf{V} = \mathbf{X}\mathbf{W}_v^T. \qquad (53)$$

For a token $\mathbf{x}_i$, the corresponding value vector is

$$\mathbf{v}_i = \mathbf{x}_i\mathbf{W}_v^T. \qquad (54)$$

When a dropout mask is applied at training epoch $t$, the perturbed value becomes

$$\mathbf{v}_{it} = (\mathbf{m}_{it} \odot \mathbf{x}_i)\mathbf{W}_v^T, \qquad (55)$$

where $\mathbf{m}_{it}$ denotes the independently sampled binary dropout mask and $\odot$ represents element-wise multiplication.

Let $\tilde{\mathbf{x}}_{it}$ denote the input token after dropout, i.e.,

$$\tilde{\mathbf{x}}_{it} = \mathbf{x}_i - (\mathbf{m}_{it} \odot \mathbf{x}_i). \qquad (56)$$

Then the deviation induced by dropout in the value space is

$$\mathbf{v}_i - \mathbf{v}_{it} = \tilde{\mathbf{x}}_{it}\mathbf{W}_v^T. \qquad (57)$$

To enforce consistency between the original and dropout-perturbed value representations, we penalize this deviation. Assuming $N_T$ stochastic training epochs with independently sampled dropout masks applied to $\mathbf{X}$, and denoting the stacked dropped components at epoch $t$ by $\tilde{\mathbf{X}}_t$, we define the value dropout regularizer as

$$J_v = \frac{1}{2N_T}\sum_{t=1}^{N_T} \left\|\tilde{\mathbf{X}}_t\mathbf{W}_v^T\right\|_F^2. \qquad (58)$$

Using the identity $\|A\|_F^2 = \text{Tr}(A^T A)$, this becomes

$$J_v = \frac{1}{2N_T} \sum_{t=1}^{N_T} \text{Tr}\left(\mathbf{W}_v \tilde{\mathbf{X}}_t^T \tilde{\mathbf{X}}_t \mathbf{W}_v^T\right). \tag{59}$$

Rearranging terms and collecting the summation yields

$$J_v = \frac{1}{2} \text{Tr}\left(\left(\frac{1}{N_T} \sum_{t=1}^{N_T} \tilde{\mathbf{X}}_t^T \tilde{\mathbf{X}}_t\right) \mathbf{W}_v^T \mathbf{W}_v\right). \tag{60}$$

Using the definition of $\mathbf{B}$ in Eq. (4), we finally obtain the compact expression

$$J_v = \frac{1}{2} \text{Tr}\left(\mathbf{B} \mathbf{W}_v^T \mathbf{W}_v\right). \tag{61}$$

### A.5. Derivation of the Value Dropout Regularizer (Attention-conditioned)

The value representation can also be obtained by

$$\mathbf{V} = \mathbf{A}\mathbf{X}\mathbf{W}_v^T. \tag{62}$$

For a token $\mathbf{x}_i$, the corresponding value vector is

$$\mathbf{v}_i = \mathbf{A}\mathbf{x}_i \mathbf{W}_v^T. \tag{63}$$

When a dropout mask is applied at training epoch $t$, the perturbed value becomes

$$\mathbf{v}_{it} = \mathbf{A}(\mathbf{m}_{it} \odot \mathbf{x}_i)\mathbf{W}_v^T, \tag{64}$$

where $\mathbf{m}_{it}$ denotes the independently sampled binary dropout mask and $\odot$ represents element-wise multiplication.

Let $\tilde{\mathbf{x}}_{it}$ denote the input token after dropout, i.e.,

$$\tilde{\mathbf{x}}_{it} = \mathbf{x}_i - (\mathbf{m}_{it} \odot \mathbf{x}_i). \tag{65}$$

Then the deviation induced by dropout in the value space is

$$\mathbf{v}_i - \mathbf{v}_{it} = \mathbf{A}\tilde{\mathbf{x}}_{it} \mathbf{W}_v^T. \tag{66}$$

To enforce consistency between the original and dropout-perturbed value representations, we penalize this deviation. Assuming $N_T$ stochastic training epochs with independently sampled dropout masks applied to $\mathbf{X}$, and denoting the stacked dropped components at epoch $t$ by $\tilde{\mathbf{X}}_t$, we define the value dropout regularizer as

$$J_v = \frac{1}{2N_T} \sum_{t=1}^{N_T} \left\|\mathbf{A}\tilde{\mathbf{X}}_t \mathbf{W}_v^T\right\|_F^2. \tag{67}$$

Using the identity $\|A\|_F^2 = \text{Tr}(A^T A)$, this becomes

$$J_v = \frac{1}{2N_T} \sum_{t=1}^{N_T} \text{Tr}\left(\mathbf{W}_v \tilde{\mathbf{X}}_t^T \mathbf{A}^T \mathbf{A} \tilde{\mathbf{X}}_t \mathbf{W}_v^T\right) \tag{68}$$

$$= \frac{1}{2N_T} \sum_{t=1}^{N_T} \text{Tr}\left(\tilde{\mathbf{X}}_t^T \mathbf{A}^T \mathbf{A} \tilde{\mathbf{X}}_t \mathbf{W}_v^T \mathbf{W}_v\right). \tag{69}$$

Rearranging terms and collecting the summation yields

$$J_v = \frac{1}{2} \text{Tr}\left(\left(\frac{1}{N_T} \sum_{t=1}^{N_T} \tilde{\mathbf{X}}_t^T \mathbf{Y} \tilde{\mathbf{X}}_t\right) \mathbf{W}_v^T \mathbf{W}_v\right). \tag{70}$$

where $\mathbf{Y} = \mathbf{A}^T \mathbf{A}$. We define

$$\psi = \frac{1}{N_T} \sum_{t=1}^{N_T} \tilde{\mathbf{X}}_t^T \mathbf{Y} \tilde{\mathbf{X}}_t. \tag{71}$$

Then,

$$J_v = \frac{1}{2} \text{Tr}\left(\psi \mathbf{W}_v^T \mathbf{W}_v\right). \tag{72}$$

#### A.5.1. Expansion of $\psi$

Expanding Eq 71,

$$\psi = \frac{1}{N_T} \sum_{t=1}^{N_T} [(\mathbf{X} - \mathbf{X}_t)^T \mathbf{Y}(\mathbf{X} - \mathbf{X}_t)] \tag{73}$$

$$= \frac{1}{N_T} \sum_{t=1}^{N_T} [\mathbf{X}^T \mathbf{Y}\mathbf{X} - \mathbf{X}^T \mathbf{Y}\mathbf{X}_t - \mathbf{X}_t^T \mathbf{Y}\mathbf{X} + \mathbf{X}_t^T \mathbf{Y}\mathbf{X}_t]. \tag{74}$$

Substituting $\mathbf{X}_t = \mathbf{M}_t \odot \mathbf{X}$,

$$\psi = \frac{1}{N_T} \sum_{t=1}^{N_T} [\mathbf{X}^T \mathbf{Y}\mathbf{X} - \mathbf{X}^T \mathbf{Y}(\mathbf{M}_t \odot \mathbf{X}) - \tag{75}$$

$$(\mathbf{M}_t \odot \mathbf{X})^T \mathbf{Y}\mathbf{X} + (\mathbf{M}_t \odot \mathbf{X})^T \mathbf{Y}(\mathbf{M}_t \odot \mathbf{X})]$$

$$= \frac{1}{N_T} \sum_{t=1}^{N_T} \left[\left((1 - \mathbf{M}_t) \odot \mathbf{X}\right)^T \mathbf{Y}\left((1 - \mathbf{M}_t) \odot \mathbf{X}\right)\right] \tag{76}$$

Following an element-wise analysis, for the $(i, j)$ entry of $\psi$, where $i, j = 1, \cdots, d$ and $a, b = 1, \cdots, n$,

$$\psi_{ij} = \frac{1}{N_T} \sum_{t=1}^{N_T} \sum_{a=1}^{n} \sum_{b=1}^{n} \left[(1 - m_{t,ai}) X_{ai} Y_{ab} (1 - m_{t,bj}) X_{bj}\right] \tag{77}$$

$$= \frac{1}{N_T} \sum_{t=1}^{N_T} \sum_{a=1}^{n} \sum_{b=1}^{n} \left[(1 - m_{t,ai})(1 - m_{t,bj}) X_{ai} Y_{ab} X_{bj}\right] \tag{78}$$

$$= \frac{1}{N_T} \sum_{t=1}^{N_T} \sum_{a=1}^{n} \sum_{b=1}^{n} ((X_{ai} Y_{ab} X_{bj})(1 - m_{t,ai} \tag{79}$$

$$- m_{t,bj} + m_{t,ai} m_{t,bj}))$$

$$= \sum_{a=1}^{n} \sum_{b=1}^{n} (X_{ai} Y_{ab} X_{bj})\left[\frac{1}{N_T} \sum_{t=1}^{N_T} (1 - m_{t,ai} \tag{80}\right.$$

$$\left. - m_{t,bj} + m_{t,ai} m_{t,bj})\right]$$

To compute the inner summation, we can apply an importance sampling Monte Carlo integration method in reverse. According to importance sampling Monte Carlo,

$$\int f(x)p(x)\,dx \approx \frac{1}{N_T}\sum_{t=1}^{N_T} f(x_t), \quad x_t \sim p(x). \tag{81}$$

Applying this to our case,

$$\frac{1}{N_T}\sum_{t=1}^{N_T}(1 + m_t^1 m_t^2 - m_t^1 - m_t^2) \tag{82}$$

$$\approx \iint (1 + m^1 m^2 - m^1 - m^2)P(m^1)P(m^2)\,dm^1 dm^2. \tag{83}$$

Since both $m^1$ and $m^2$ are binary, the integral has only four possible values:

$$\iint (1+m^1 m^2 - m^1 - m^2) \times P(m^1)P(m^2)\,dm^1\,dm^2 \tag{84}$$

$$= S_1 + S_2 + S_3 + S_4 = p^2, \tag{85}$$

$$P(m^1) = (1-p)^{m^1} p^{(1-m^1)}, \tag{86}$$

$$P(m^2) = (1-p)^{m^2} p^{(1-m^2)}, \tag{87}$$

$$S_1 = (1 + 1 - 1 - 1)(1-p)^2, \tag{88}$$

$$S_2 = (1 + 0 - 1 - 0)p(1-p), \tag{89}$$

$$S_3 = (1 + 0 - 0 - 1)(1-p)p, \tag{90}$$

$$S_4 = (1 + 0 - 0 - 0)p^2. \tag{91}$$

Therefore,

$$\psi \approx (\mathbf{X}^T \mathbf{Y} \mathbf{X})p^2. \tag{92}$$

## A.6. Derivation of the Feedforward Network Dropout Regularizer

We derive the regularizer for a single linear transformation; the extension to multi-layer FFNs follows analogously. Consider the position-wise feedforward layer

$$\mathbf{H} = \mathbf{X}\mathbf{W}_{\text{ff}}^T, \tag{93}$$

where $\mathbf{W}_{\text{ff}}$ denotes the feedforward weight matrix.

For a token $\mathbf{x}_i$, the corresponding output is

$$\mathbf{H}_i = \mathbf{x}_i \mathbf{W}_{\text{ff}}^T. \tag{94}$$

When a dropout mask $\mathbf{m}_{it}$ is applied at training epoch $t$, the perturbed output becomes

$$\mathbf{H}_{it} = (\mathbf{m}_{it} \odot \mathbf{x}_i)\mathbf{W}_{\text{ff}}^T. \tag{95}$$

The discrepancy between the original and perturbed outputs is therefore

$$\mathbf{H}_i - \mathbf{H}_{it} = \tilde{\mathbf{x}}_{it}\mathbf{W}_{\text{ff}}^T, \tag{96}$$

where, $\tilde{\mathbf{x}}_{it} = \mathbf{x}_i - (\mathbf{m}_{it} \odot \mathbf{x}_i)$.

Assuming $N_T$ stochastic training epochs with independently sampled dropout masks applied to $\mathbf{X}$, and denoting the stacked dropped components at epoch $t$ by $\tilde{\mathbf{X}}_t$, we define the feedforward dropout regularizer as

$$J_{\text{ff}} = \frac{1}{2N_T}\sum_{t=1}^{N_T} \left\| \tilde{\mathbf{X}}_t \mathbf{W}_{\text{ff}}^T \right\|_F^2. \tag{97}$$

Using $\|A\|_F^2 = \text{Tr}(A^T A)$, we obtain

$$J_{\text{ff}} = \frac{1}{2N_T}\sum_{t=1}^{N_T} \text{Tr}\left(\mathbf{W}_{\text{ff}} \tilde{\mathbf{X}}_t^T \tilde{\mathbf{X}}_t \mathbf{W}_{\text{ff}}^T\right). \tag{98}$$

Rearranging terms yields

$$J_{\text{ff}} = \frac{1}{2}\text{Tr}\left(\left(\frac{1}{N_T}\sum_{t=1}^{N_T} \tilde{\mathbf{X}}_t^T \tilde{\mathbf{X}}_t\right) \mathbf{W}_{\text{ff}}^T \mathbf{W}_{\text{ff}}\right). \tag{99}$$

Using the definition of $\mathbf{B}$ in Eq. (4), the regularizer admits the compact form

$$J_{\text{ff}} = \frac{1}{2}\text{Tr}\left(\mathbf{B}\mathbf{W}_{\text{ff}}^T \mathbf{W}_{\text{ff}}\right). \tag{100}$$

## B. Relationship Between the Proposed Explicit Dropout Regularizer and the Prior Explicit Dropout Regularization [1]

In this section, we show that the proposed explicit dropout regularizer can be decomposed into the previously derived explicit dropout regularizer [1] plus an additional structured attention-dependent term. This provides theoretical insight into how our formulation generalizes earlier explicit dropout formulations.

### B.1. Preliminaries

For completeness and clarity, we rewrite the explicit dropout regularizer of [1] for the special case of a single linear layer. Consider a supervised learning problem with a batch of inputs $\mathbf{X} = [\mathbf{x}_1, \mathbf{x}_2, \ldots, \mathbf{x}_N]$ where each $\mathbf{x}_i \in \mathbb{R}^{1 \times d}$, targets $y_i$, and a linear layer $f(\mathbf{X}; \mathbf{W}) = \mathbf{X}\mathbf{W}^T$ with weights $\mathbf{W} \in \mathbb{R}^{d_1 \times d}$. For a mini-batch $\mathbf{X} \in \mathbb{R}^{N \times d}$, the empirical task loss is

$$J_{\text{task}}(\mathbf{X}, \mathbf{W}) = \frac{1}{n}\sum_{r=1}^{N} \ell(f(\mathbf{x}_i; W), y_i), \tag{101}$$

where $\mathbf{x}_r$ is the $i$-th input of a mini-batch.

Feature-wise dropout with probability $p$ multiplies each feature by a Bernoulli mask. Prior work [1] shows that the expected dropout objective can be written as

$$J(\mathbf{X}, \mathbf{W}) = J_{\text{task}}(\mathbf{X}, \mathbf{W}) + \widehat{R}(\mathbf{X}, \mathbf{W}), \tag{102}$$

where the explicit regularizer for a single linear layer is

$$\widehat{R}(\mathbf{X}, \mathbf{W}) = \frac{p}{1-p}\sum_{j=1}^{d} \|\mathbf{W}_j\|_2^2 \left(\frac{1}{n}\sum_{r=1}^{N} X_{rj}^2\right) \tag{103}$$

$$= \frac{p}{1-p}\sum_{j=1}^{d} \hat{\sigma}_j^2 \|\mathbf{W}_j\|_2^2, \tag{104}$$

with $\hat{\sigma}_j^2$ being the empirical second moment of feature $j$.

## B.2. Proposed Explicit Dropout Regularizer

Our proposed formulation introduces a structured explicit dropout regularizer for a single linear layer. The final training objective is

$$J(\mathbf{X}, \mathbf{W}) = J_{\text{task}}(\mathbf{X}, \mathbf{W}) + R(\mathbf{X}, \mathbf{W}), \quad (105)$$

where,

$$R(\mathbf{X}, \mathbf{W}) = \frac{p^2}{2n} \sum_{r=1}^{N} \sum_{k=1}^{d_1} \sum_{i=1}^{d} \sum_{j=1}^{d} X_{ri} X_{rj} W_{ki} W_{kj}. \quad (106)$$

Here $n$ is the batch size, $W \in \mathbb{R}^{d_1 \times d}$ is the weights matrix of the linear layer, and $p$ is the dropout probability.

Each term in the summation corresponds to a contribution from a specific pair of input features $(i, j)$ and output unit $k$, capturing both diagonal (feature-wise) and off-diagonal (cross-feature) interactions. The diagonal component recovers the prior explicit dropout regularizer [1], while the off-diagonal terms introduce additional structured regularization dependent on feature covariances.

## B.3. Decomposition of the Proposed Regularizer

We now show that the proposed regularizer can be written as

$$R(\mathbf{X}, \mathbf{W}) = \widehat{R}(\mathbf{X}, \mathbf{W}) + \Delta R(\mathbf{X}, \mathbf{W}), \quad (107)$$

$$= \frac{p^2}{2n} \sum_{r=1}^{N} \sum_{k=1}^{d_1} \left( \sum_{i=j}^{d} X_{ri}^2 W_{ki}^2 + \sum_{i \neq j}^{d} X_{ri} X_{rj} W_{ki} W_{kj} \right). \quad (108)$$

*Diagonal Term ($i = j$).* The first term corresponds to the classical explicit dropout regularizer structure:

$$R_{\text{diag}}(\mathbf{X}, \mathbf{W}) = \frac{p^2}{2n} \sum_{r=1}^{N} \sum_{k=1}^{d_1} \sum_{j=1}^{d} X_{rj}^2 W_{kj}^2. \quad (109)$$

Rearranging the sums gives

$$R_{\text{diag}}(\mathbf{X}, \mathbf{W}) = \frac{p^2}{2} \sum_{j=1}^{d} \sum_{k=1}^{d_1} W_{kj}^2 \left( \frac{1}{N} \sum_{r=1}^{N} X_{rj}^2 \right) \quad (110)$$

$$= \frac{p^2}{2} \sum_{j=1}^{d} \|\mathbf{W}_j\|_2^2 \hat{\sigma}_j^2.$$

Hence, the diagonal term is proportional to the prior explicit dropout regularizer:

$$R_{\text{diag}}(\mathbf{X}, \mathbf{W}) = \underbrace{\frac{p(1-p)}{2}}_{\alpha} \widehat{R}(\mathbf{X}, \mathbf{W}), \quad (111)$$

where $\alpha$ is a scaling factor determined by the constants in the two formulations.

*Off-Diagonal Term ($i \neq j$).* The remaining component captures cross-feature interactions:

$$R_{\text{cross}}(\mathbf{X}, \mathbf{W}) = \frac{p^2}{2n} \sum_{r=1}^{N} \sum_{k=1}^{d_1} \sum_{i \neq j} X_{ri} X_{rj} W_{ki} W_{kj}. \quad (112)$$

*Final Decomposition.* Combining both terms yields

$$R(\mathbf{X}, \mathbf{W}) = \alpha \widehat{R}(\mathbf{X}, \mathbf{W}) + R_{\text{cross}}(\mathbf{X}, \mathbf{W}). \quad (113)$$

This shows that the proposed regularizer extends the classical explicit dropout regularizer by incorporating cross-feature covariance terms, which capture interactions between different input dimensions rather than only feature-wise magnitudes. These additional terms provide a finer control over generalization by accounting for correlations in the input data.

## CRediT authorship contribution statement

**Vidhi Agrawal:** Formal analysis, Investigation, Methodology, Software, Validation, Visulaization, Writing – original draft. **Illia Oleksiienko:** Formal analysis, Supervision, Writing – review and editing. **Alexandros Iosifidis:** Conceptualization, Supervision, Writing – review and editing.